%% file: main.tex
\documentclass{article} 
\usepackage{iclr2027_conference,times}

\input{math_commands.tex}

\usepackage{inconsolata}
\PassOptionsToPackage{table}{xcolor}
\usepackage[dvipsnames,table]{xcolor}
\colorlet{themecolor}{MidnightBlue!90!black}
\colorlet{highlight}{Orchid!90!black}
\usepackage[colorlinks,linkcolor=themecolor,citecolor=themecolor,urlcolor=themecolor,pagebackref=true]{hyperref}
\usepackage[capitalise,nameinlink]{cleveref}
\renewcommand*{\backrefalt}[4]{\mbox{(cited on \ifnum#1=1 page~\else pages~\fi #2)}}
\bibpunct{\color{themecolor}(}{\color{themecolor})}{;}{a}{,}{,}

\usepackage{titletoc}
\usepackage[most]{tcolorbox}
\newtcolorbox{highlightbox}[0]{colback=themecolor!10,left=6pt,right=6pt,top=4pt,bottom=4pt,boxrule=0pt,opacityframe=0,arc=0pt,breakable}

\usepackage{url}
\usepackage[utf8]{inputenc}
\usepackage{booktabs}
\usepackage{tabularx}
\usepackage{amsmath}
\usepackage{caption}
\usepackage{graphicx}

\title{Accelerating SO(2)-Equivariant Materials \\ Models through Low-Precision Emulation}

\author{Denghui Lu, Alexander Maeder, Alexandros Nikolaos Ziogas, \\
\textbf{Mathieu Luisier \& Manasa Kaniselvan} \\
D-ITET, ETH Zurich, Zurich, Switzerland \\
\texttt{\{denghuilu,almaeder,alziogas,mluisier,mkaniselvan\}@iis.ee.ethz.ch}
}

\iclrfinalcopy 
\begin{document}

\maketitle
\thispagestyle{plain}
\lhead{}
\chead{Accelerating SO(2)-Equivariant Materials Models through Low-Precision Emulation}

\begin{abstract}
Rotationally equivariant architectures improve the physical fidelity of machine-learned (ML) universal materials models based on graph neural networks (GNNs). However, their scalability is limited by the use of non-trivial, high-precision tensor operations during both training and inference. Adopting the reduced-precision formats of matrix multiplication units (MMUs), e.g., FP8 instead of FP32, provides massive increases in throughput, up to 30$\times$ (60$\times$) on NVIDIA Hopper (Blackwell) GPUs, but may lead to drastic loss of accuracy. Here, we introduce tailored `emulated' low-precision operations under the integer Ozaki scheme to accelerate the `SO(2)-equivariant' processing layers of GNNs, which dominate the runtime of (1) many state-of-the-art ML interatomic potentials (MLIPs) and (2) most universal ML Hamiltonian (MLH) models. Our framework, \textit{FlashSO(2)}, outperforms TF32, the most accurate native low-precision MMU format, in throughput, accuracy, and memory footprint at once. 
At the $\mathrm{SO}(2)$ layer level, it reaches $40\times$ ($7\times$) the throughput of MALOQ's (UMA's) reference implementation, an MLH (MLIP) model,  sustaining up to 236 (232) effective Tflop/s on an NVIDIA GH200, $3.5\times$ the FP32 peak. 
At the model level, \textit{FlashSO(2)} helps train MALOQ $3.9\times$ faster than its pure-FP32 version while reducing peak memory by $58\%$ at no loss in accuracy. In  molecular dynamics simulations with UMA, the speedup factor is $3.3\times$, and the peak memory is reduced by $51\%$. \textit{FlashSO(2)} thus pushes the speed-accuracy Pareto frontier of equivariant materials models beyond the native low-precision formats and substantially lowers the resources required to exploit their scaling laws.

\end{abstract}

\section{Introduction}

Universal machine-learned (ML) models based on graph neural networks (GNNs) are revolutionizing computational materials science by bringing \textit{ab initio} calculations of key properties such as energies, forces, or electronic structures to previously unfeasible atomic scales and elemental diversity. Beyond the standard ML interatomic potentials (MLIPs) \citep{uma, esen, mace, pet}, which drive molecular dynamics (MD) engines, the underlying architectures have been adapted to effectively learn higher-rank matrix-valued properties including Hamiltonians \citep{helm, wanet, deeph_pack, qhflow, hamgnn} and Hessians \citep{hip}. Improvements in representation \citep{e3nn} and engineering of inductive biases \citep{nequip} have played a pivotal role in bringing ML materials models close to a suitable accuracy, at the level of density-functional theory (DFT), for downstream scientific applications across different fields. 

\begin{figure}[t]
  \centering
  \includegraphics[width=\linewidth]{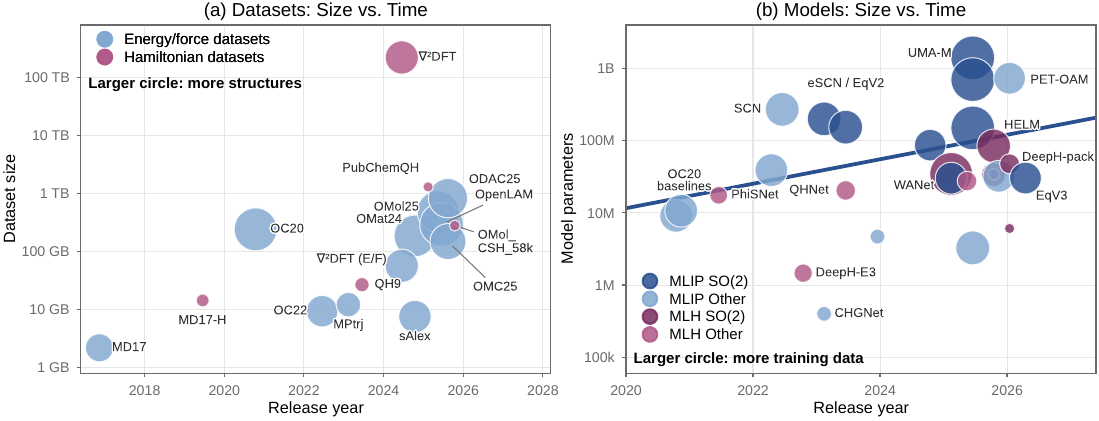}
  \caption{Historical trends in ML materials models and data scaling. (a) Training datasets by release year and size on disk. 
  (b) MLIP (blue) and MLH (pink) models by release year and parameter count.
  }
\label{fig:dataset_bubble_plot_gb}
\end{figure}

Progress in universal materials models is now increasingly driven by aggressive scaling across both data volume and model capacity. Today's largest atomic structure datasets consist of on the order of 100M structures (\textbf{Fig.~\ref{fig:dataset_bubble_plot_gb}}(a)). For a given set of structures, going from their simple scalar (energies) to their vector (forces) properties, up to high-rank operators (Hamiltonians, density matrices), increases the volume of training data available by several orders of magnitude. MLIP and ML Hamiltonian (MLH) prediction models are currently at typical scales of 110M parameters (\textbf{Fig.~\ref{fig:dataset_bubble_plot_gb}}(b)), and scaling laws indicate that their performance will continue to enhance with further increases in model size \citep{uma}. Improvement along these axes thus presents the most direct avenue to further push their capabilities. 

Increasing the number of learnable parameters and training on terabytes of DFT datasets pose significant computational challenges that can be addressed at the theory, algorithm, and/or hardware level. Initially, the scalability of the ML architectures was tackled through mathematical simplification of their tensor operations. \citet{so2} introduced a rotationally-equivariant layer, analogous to full Clebsch--Gordan Tensor Products (CGTPs) \citep{e3nn}. By reducing the computational complexity from $O(l_{max}^6)$ to $O(l_{max}^3)$, this innovation enabled scaling to higher maximum spherical degree ($l_{max}$) and to a larger number of learnable parameters. These `$\mathrm{SO}(2)$-equivariant layers' were adopted in several MLIPs such as eSEN \citep{esen}, EquiformerV2 \citep{equiformerv2}, UMA \citep{uma}, and DPA4 \citep{dpa4}. The ability to process higher spherical embedding ranks also unlocks efficient treatment of material properties involving larger $l_{max}$, e.g., electronic structures. SO(2) layers thus currently power nearly all scalable MLH models, for example DeepH-pack \citep{deeph_pack}, WANet \citep{wanet}, HELM \citep{helm}, and MALOQ \citep{maloq}.

$\mathrm{SO}(2)$ layers are typically executed in FP32 precision. Despite their reduced arithmetic complexity compared to CGTPs, they still account for more than 80\% of each inference step (\textbf{Fig.}~\ref{fig:uma_results}(a)), \textit{leaving the hardware level as the primary remaining lever to further improve scalability.} 
On the other hand, mainstream machine learning and large language models (LLMs) rely on lower-precision operations, e.g., FP16, BF16, or FP8 \citep{the_quantization_paper, mixed_precision}, performed on matrix multiplication units (MMUs), e.g., tensor or matrix cores. On NVIDIA Hopper GPUs, these MMUs offer 7$\times$ (TF32) to 30$\times$ (FP8) the peak throughput of general-purpose FP32 units. 
LLMs can exploit them because their linear and attention layers tolerate the rounding errors of reduced precision \citep{fp8, deepseekv3, fa3}.
Symmetry-constrained GNNs lack this robustness. 
Although some MLIP and MLH models pretrain in BF16 to save compute time, low precision formats remain unsuitable for fine-tuning and inference as they introduce random error under rotations. Notably, BF16 alone degrades UMA's accuracy by 20\% to 50\% \citep{uma}. 
Symmetry-constrained ML materials models, therefore, spend most of their compute time in FP32 or even FP64 general-purpose units, and leave the MMUs idle.

Here, we address this gap through custom \textit{precision emulation}, notably the Ozaki scheme \citep{ozaki2012}, which reconstructs high-precision matrix products from exact low-precision `slice' products. Specifically, we introduce \textit{FlashSO(2)}, a software-hardware co-designed implementation of the $\mathrm{SO}(2)$ layer with the integer Ozaki emulation \citep{ootomo2024}, where the slice count obeys equivariant constraints, the scaling follows expected tensor value distributions, while slicing and reconstruction are fused with the surrounding kernels. 
As a result, the $\mathrm{SO}(2)$ layer can run on INT8 MMUs within a $10^{-4}$ equivariance tolerance, and \textit{FlashSO(2)} simultaneously demonstrates better accuracy, higher throughput, and a smaller memory footprint than the native TF32 format.
At the level of the $\mathrm{SO}(2)$ layer, it reaches $40\times$ the throughput of the reference implementation in the MALOQ MLH architecture ($l_{max}$ = 8) and $7\times$ in the UMA-Large (UMA-L) MLIP ($l_{max}$ = 6), sustaining 236 and 232 effective Tflop/s on an NVIDIA GH200 superchip, $3.5\times$ the FP32 peak. 
As the layer dominates each step,  MALOQ trains $3.9\times$ faster end-to-end than FP32 at 58\% lower peak memory. UMA-Medium (UMA-M) runs training (inference) steps $2.6\times$ ($3.3\times$) faster, at 41\% (51\%) lower peak memory.
A memory-lean mode called `MiniMem' further cuts peak memory by 81\% and extends single-GPU inference from 5.3k atoms (85.2k edges) in FP32 to 25.6k (410k edges), all while accelerating training by 1.9$\times$ compared to FP32.

\section{Related work}

\paragraph{Equivariant MLIP and MLH architectures}
\label{subsec:mlh_and_mlip}

Mainstream GNN-based MLIP and MLH models take advantage of spatial locality and interatomic interaction decay with increased input structure sizes, i.e., number of atoms. One of the main advances in reaching realistic physical fidelity has been the integration of rotational symmetry into the backbone architectures of these models \citep{nequip, mace, deeph_pack}. Doing so restricts the learnable space to physically valid representations and enforces physical properties that standard loss functions, e.g., mean absolute error, may not capture. We refer to \cite{so2} for the relevant mathematical background behind the transformation of CGTPs into more computationally efficient $\mathrm{SO}(2)$ convolutions, which serve as the core message-passing (MP) engine across several universal MLIP and MLH architectures.

An important difference between the equivariant backbone of energy/force and Hamiltonian matrix prediction resides in the maximum angular momentum, $l_{max}$. MLIPs treat $l_{max}$ as a hyperparameter and increase it to incorporate higher-frequency intermediate information into the node/edge embeddings, which are eventually filtered out within the output head to predict scalar (energy $E$) and vector (force $F$) quantities. Meanwhile, MLH models set $l_{max}$ to match the dimensionality of the data: For a given atomic orbital basis with highest orbital degree $l_1$, the interaction which produces the maximum output degree is $l_1+l_1$ \citep{schnorb}. For example, datasets containing $d$-orbitals ($l=2$: MD17 \citep{schnorb}, QH9 \citep{qh9}, or $\nabla^2$DFT \citep{nabla2dft}) require $l_{max}=4$, those containing $f$-orbitals ($l=3$: OMol\_CSH\_58k \citep{helm}, PubChemQH \citep{wanet}), $l_{max}=6$. Special datasets including complex compounds such as lanthanides, as encountered in the full OMol25 electronic structure data \citep{omol25}, require $l_{max}=8$ due to the presence of $g$-orbitals.

\paragraph{Acceleration of ML materials models}

Recent efforts to scale MLIPs to larger system sizes and longer timescales have reduced both training and inference times at the software and system level. 
At the software level, EquiformerV3 \citep{equiformerv3} fused the permutation matrices of its $\mathrm{SO}(2)$ linear layers into the Wigner-D rotations and leveraged PyTorch's Just-In-Time compiler (torch.compile) 
\citep{ansel2024pytorch2} to eliminate redundant operations, yielding a 1.75$\times$ training speedup over EquiformerV2 \citep{equiformerv2}. DPA4 made conservative force training compatible with torch.compile, cutting its training time per step threefold \citep{dpa4}. 
At the system level, \citet{esen_big} ran the eSEN-omol backbone through a distributed inference framework that spreads fully solvated enzymes of 32k to 54k atoms over 8 to 32 GPUs, reaching millions of MD steps per day and 1 $\mu$s of aggregated simulation time. 

Emerging $\mathrm{SO}(2)$-convolution-based MLH architectures have also started to focus on scalability across structure size, orbital basis, and elemental diversity. DeepH-pack demonstrated inference on 4,096-atom amorphous Silicon systems \citep{deeph_pack}. WANet integrated custom loss functions to improve scalability to elements with larger basis sets and requiring higher spherical degree \citep{wanet}. HELM leveraged advanced batching schemes to train on the OMol\_CSH\_58k universal Hamiltonian dataset \citep{helm}. MALOQ introduced graph partitioning approaches to tackle Hamiltonians representing systems consisting of 10k+ atoms, comparable to experimentally fabricated electronic devices \citep{maloq}. However, the core $\mathrm{SO}(2)$ layer bottleneck has yet to be sped up.
We note that many kernel libraries for equivariant networks, for example cuEquivariance \citep{cuequivariance}, OpenEquivariance \citep{openequivariance}, and FlashTP \citep{flashtp}, fuse the CGTPs of $\mathrm{SO}(3)$ architectures such as MACE~\citep{mace} and NequIP~\citep{nequip}. These methods do not cover the $\mathrm{SO}(2)$ convolution, whose cost lies in dense matrix products per spherical order rather than in sparse coupling contractions.

\paragraph{High-precision matrix product on low-precision units}
Precision emulation, notably the Ozaki scheme \citep{ozaki2012}, computes a high-precision matrix product on low-precision MMUs by splitting each operand into a sum of low-precision slices.
For example, the integer form of the Ozaki scheme \citep{ootomo2024} uses INT8 slices with INT32 accumulation, which makes the partial products exact on the INT8 MMUs of the GPU. NVIDIA has since adopted this form for FP64 emulation in its cuBLAS library on Blackwell \citep{cublas_emulation}. EmuGEMM \citep{emugemm} further removes the slicing and reconstruction round trips with fused kernels.

\section{Methods \& Implementation}

This section introduces the two ingredients of \textit{FlashSO(2)} and their combination. In \textbf{Section}~\ref{subsec:emulation} we provide relevant background on precision emulation, which computes a high-precision matrix product from INT8 products, together with the block scaling strategy we adopt. \textbf{Section}~\ref{subsec:pipeline} presents the implementation of the $\mathrm{SO}(2)$ layer, distinguishing between the matrix products, which carry most of the arithmetic and the steps that mainly move data.  \textbf{Section}~\ref{subsec:flashso2} describes how \textit{FlashSO(2)} builds emulation into this layer.

\begin{figure}[t]
  \centering
  \includegraphics[width=\linewidth]{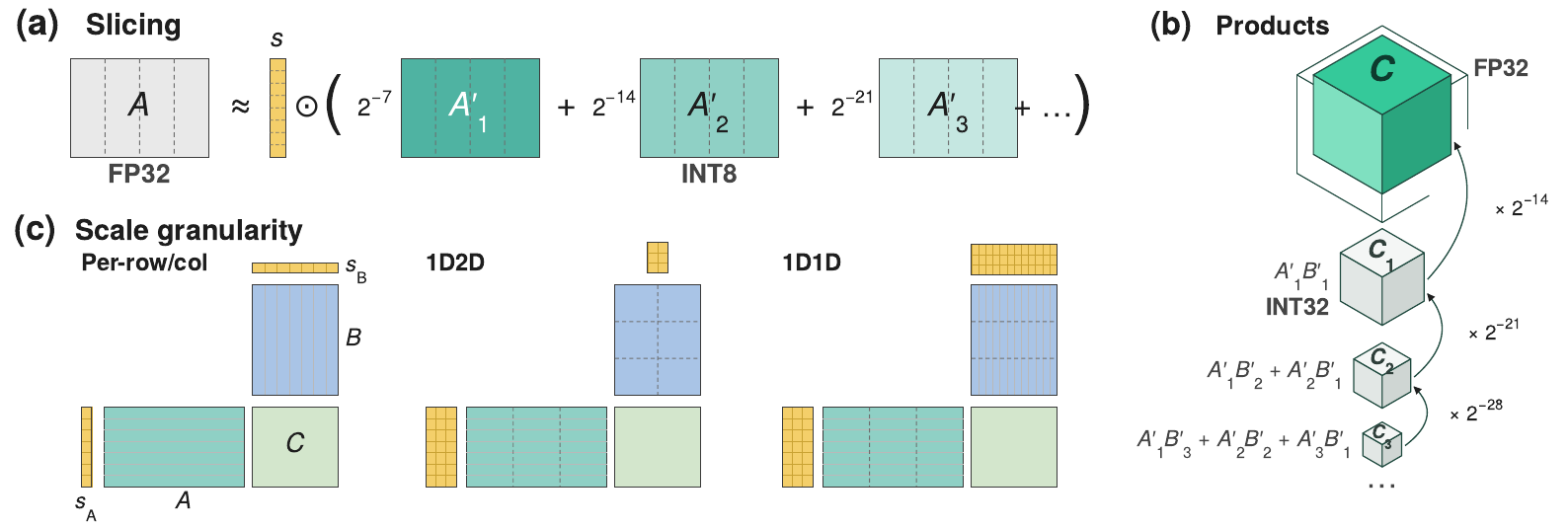}
  \caption{\textbf{Integer emulation with block scales.}
    (a)~An FP32 matrix is split into $p$ INT8 slices.
    (b)~The slice products $\mathbf{A}'_i\mathbf{B}'_j$ with $i+j \leq p+1$ are accumulated exactly in INT32 and summed in FP32 with weights $2^{-7(i+j)}$.
    (c)~Block scaling refines the scales (yellow) from one per row of $A$ and column of $B$ in the Ozaki scheme to one per 128 entries of a row (1-D) or per $128\times128$ tile (2-D).}
\label{fig:emulation}
\end{figure}

\subsection{Integer Emulation with block scaling}
\label{subsec:emulation}

\textbf{Figure \ref{fig:emulation}} illustrates the integer Ozaki scheme \citep{ootomo2024} with block scaling for a single matrix product $\mathbf{C} = \mathbf{A}\mathbf{B}$, where $\mathbf{A} \in \mathbb{R}^{M \times K}$ and $\mathbf{B} \in \mathbb{R}^{K \times N}$ are given in FP32.
The three steps below follow panels (a)--(c) of the figure. Steps (1) and (2) are the Ozaki scheme itself, while step (3) covers the block scaling.

(1) Scaling and slicing:
INT8 MMUs multiply integers, meaning that the FP32 entries must be first converted to fixed point. Specifically, each row of $\mathbf{A}$ (and each column of $\mathbf{B}$) is divided by the power of two just above its largest entry, giving the whole row a shared scale.
Each scaled entry is then cut into $p$ INT8 slices (\textbf{Fig.}~\ref{fig:emulation}(a)). The entry is the sum of its slices weighted by their place values $i$ with $2^{-7i}$.
With $p$ slices, a setting we refer to here as `Emu{$p$}', an operand can theoretically keep $7p$ bits accuracy. Notably, Emu$2$ keeps 14 bits, more than the 10 mantissa bits of TF32.

(2) Products and reconstruction:
After slicing, the product of two operands expands into $p^2$ pairs of slice products $\mathbf{A}'_i\mathbf{B}'_j$, one per pair of slices, each carrying the weight $2^{-7(i+j)}$ (\textbf{Fig.}~\ref{fig:emulation}(b)).
In practice, we only keep the products with $i + j \leq p + 1$. The result is reassembled as
\begin{equation}
    \mathbf{A}\mathbf{B} \approx \mathrm{diag}(\boldsymbol{\mu}) \Big[\, {\textstyle\sum_{i+j \leq p+1}} \, 2^{-7(i+j)} \mathbf{A}'_i \mathbf{B}'_j \,\Big] \, \mathrm{diag}(\boldsymbol{\nu}),
\label{eq:ozaki}
\end{equation}
where diag($\mu$) and diag($\nu$) restore the row and column scales of step (1).
Each slice product $\mathbf{A}'_i\mathbf{B}'_j$ is computed on INT8 MMUs with INT32 accumulation, which multiply and add integers exactly, avoiding the introduction of rounding errors. 
According to \textbf{Eq.}~\ref{eq:ozaki}, Emu$1$--Emu$3$ cost 1, 3, and 6 INT8 MMU products, respectively. The slice count, therefore, sets precision and cost in much finer steps than native formats offer.

(3) Block scaling:
The slices fix the number of bits kept by an operand, 14 for Emu$2$. The scales determine how well each entry uses them. With one scale per row, the bits are aligned to the largest entry of the row, and every factor of two below it costs one bit. An entry $2^{10} \approx 1000\times$ smaller than the row maximum thus keeps only 4 bits of the 14 bits, fewer than the 7 mantissa bits of BF16.
Block scaling \citep{deepseekv3} instead gives every block of 128 consecutive entries along $K$ its own scale (\textbf{Fig.}~\ref{fig:emulation}(c)).

\subsection{SO(2) layer pipeline}
\label{subsec:pipeline}

To eliminate the computational cost of full CGTPs, $\mathrm{SO}(2)$-convolution architectures project 3D rotational interactions onto an aligned 2D reference frame along the interatomic edge vectors. At the beginning of each forward pass, for each directed edge $(i, j)$, we precompute an origin-aligned Cartesian rotation matrix $g_{ij}$ that rotates the unit edge vector $\hat{\mathbf{r}}_{ij}$ to point along the global $\hat{\mathbf{z}}$-axis:
\(
g_{ij} \cdot \hat{\mathbf{r}}_{ij} =  \hat{\mathbf{z}}
\), and store the corresponding Wigner-$D$ matrix $\mathbf{D}(g_{ij})$. Then, each $\mathrm{SO}(2)$ layer performs the following set of operations, which are summarized schematically in \textbf{Fig.~\ref{fig:flashso2}}(a).

\textbf{Rotation to a common axis (S1):}
Features $\phi^{(l)}$, consisting of multichannel spherical harmonic coefficients from rank $0...l_{max}$ along each edge, are rotated to the global $\hat{\mathbf{z}}$-axis using the Wigner-$D$ matrices. These matrices consist of block diagonal components (\textbf{Fig.~\ref{fig:flashso2}}(a)). If $\mathbf{D}^{(l)}(g_{ij})$ is the Wigner-D matrix of rank $l$ corresponding to $g_{ij}$, the rotation of $\phi^{(l)}$ yields
\(
\tilde{\phi}^{(l)} = \mathbf{D}^{(l)}(g_{ij}) \phi^{(l)}
\).

\textbf{$\mathbf{SO}(2)$ Conv 1:} To optimize computational execution, the rotated features first undergo a layout transformation, where they are permuted from a standard `$l$-major' memory layout to an `$m$-major', i.e. the feature entries are grouped by their $m$ index. The $\mathrm{SO}(2)$ layer operation then transforms each $m$-subspace through a separate linear layer, performing up to $l_{max}+1$ GEMM operations. 

\textbf{Gated non-linear activation (S2):} A gated activation layer is typically introduced between convolutions. In this step, the purely scalar components ($l=0$) of the intermediate message undergo a standard non-linear activation $\sigma_{\text{scalar}}$ (e.g., SiLU).
Doing the same to higher-order tensor components ($l > 0$) would break equivariance. Instead, a gating function $\sigma_{\text{gate}}$ is used to scale all the $m$-components within each feature.

\textbf{$\mathbf{SO}(2)$ Conv 2:} A second $\mathrm{SO}(2)$ convolution layer operates analogously to the first one, followed by a layout transformation back to $l$-major.

\textbf{Rotation back to the global frame (S3):} The resulting messages $m_{ij}^l$ are rotated back to the global reference frame using the inverse Wigner-$D$ transformation:
\(
\mathbf{\hat{m}}_{ij}^{(l)} = \mathbf{D}^{(l)}(g_{ij}^{-1}) m_{ij}^l
\). Node updates are obtained by aggregating these messages: $\mathbf{h}_i' = \mathbf{h}_i + \sum_{j \in \mathcal{N}(i)} \mathbf{\hat{m}}_{ij}$, where $\mathcal{N}(i)$ represents the neighborhood of node $i$ within a specified cutoff.

\subsection{Optimizing precision emulation for SO(2) layers}
\label{subsec:flashso2}

The $\mathrm{SO}(2)$ layer of \textbf{Section}~\ref{subsec:pipeline} proceeds in a chain of memory-bound stages.
The time spent in this layer depends largely on two things: memory traffic and matrix products. \textit{FlashSO(2)} addresses both with the following strategies.

\begin{figure}[t]
  \centering
  \includegraphics[width=\linewidth]{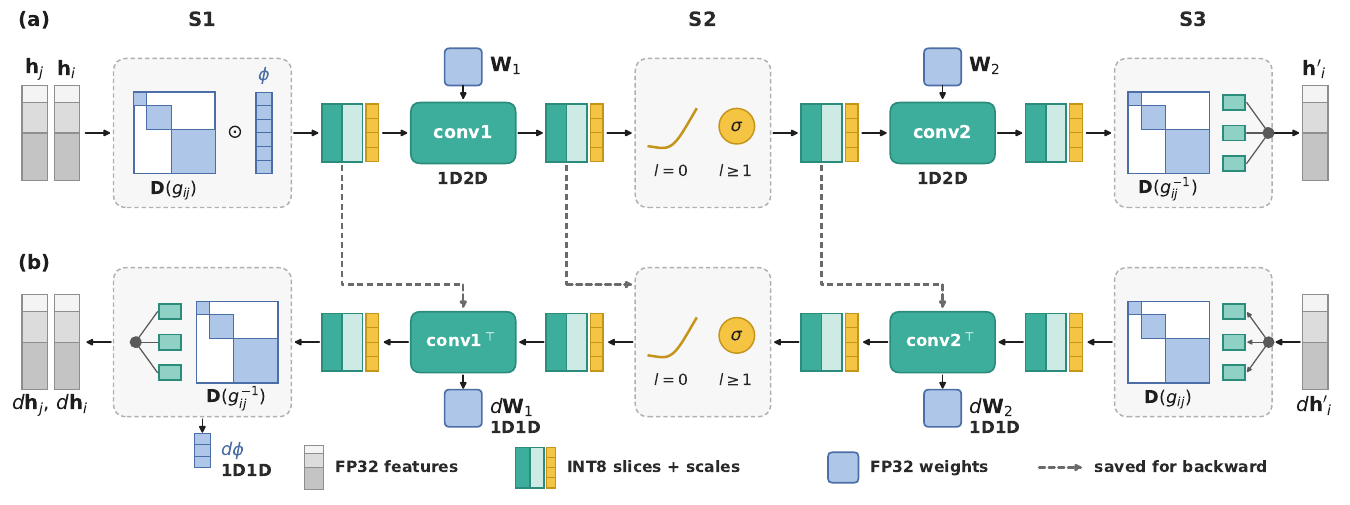}
  \caption{\textbf{FlashSO(2): the $\mathrm{SO}(2)$-convolution block and its fused implementation.}
    (a)~Forward pass; each dashed box S1 to S3 is one fused kernel: Only INT8 slices and their scales pass between kernels.
    (b)~Backward pass (read from right to left); each kernel is placed under its forward counterpart. Dashed arrows mark the tensors saved in the forward pass.}
\label{fig:flashso2}
\end{figure}

(1) Raising arithmetic intensity.
\textit{FlashSO(2)} fuses everything around the GEMMs of the forward pass into three kernels, the dashed boxes S1 to S3 in \textbf{Fig.~\ref{fig:flashso2}}(a). Inside each kernel, the intermediate results stay in registers instead of being transferred to off-chip memory in between the separate PyTorch kernels of the reference implementation. 
Between kernels, only INT8 slices and their scales are written. 
S1 gathers, rotates, and weights the edge features. S2 applies the activation and the gate, computed in its first pass over the orders $m$ and never stored.
S3 rotates the messages back and sums them into the nodes without a message buffer. 
The backward pass, \textbf{Fig.~\ref{fig:flashso2}}(b), mirrors this. Each convolution backward runs two GEMMs per order $m$, the activation gradients continue in INT8, and the weight gradients $d\mathbf{W}$ are returned in FP32. 
S1 also returns $d\phi$. This operation only needs the INT8 slices, scales, and gate logits (dashed arrows) from the forward pass, so again, no FP32 activation is stored. 

(2) Emulating high-precision products in INT8 MMUs.
With the chain of operations in the $\mathrm{SO}(2)$ layer fused, the GEMMs (conv1/2 boxes in \textbf{Fig.~\ref{fig:flashso2}}) dominate the execution time; \textit{FlashSO(2)} runs them on the INT8 MMUs with the block-scaled emulation of \textbf{Section}~\ref{subsec:emulation}.
Block scaling is needed here for two reasons. The activations of the $\mathrm{SO}(2)$ layer spread over orders of magnitude within a row, so with one scale per row, small entries keep few bits. The weight gradient, in turn, sums over the edge dimension, which in equivariant models could exceed 65,536 ($2^{16}$), at which point an INT32 sum over a whole row could overflow. 
One scale per block of 128 entries (\textbf{Fig.~\ref{fig:emulation}}(c)) fixes both issues.
The two operands are not equally demanding: The activations and their gradients carry one scale per 128 entries of a row (1-D), whereas the evenly distributed weights need only one per $128\times128$ tile (2-D). The forward GEMM and the activation gradient are thus 1-D$\times$2-D products, the weight gradient 1-D$\times$1-D.

(3) Compatibility with system-level optimizations.
As illustrated in \textbf{Fig.}~\ref{fig:flashso2}, \textit{FlashSO(2)} is a mixed-precision implementation. The weights enter and leave the $\mathrm{SO}(2)$ layer in FP32, the INT8 slices exist only inside it, and every other module of the model, including the optimizer, runs in FP32.
Therefore, \textit{FlashSO(2)} is completely orthogonal to standard system-level optimizations, which apply on top of it unchanged.
Torch.compile \citep{ansel2024pytorch2} still fuses the small operations outside the layer and removes their Python overhead, and CUDA graphs still remove the launch latency of the remaining kernels. 
To replace standard activation check-pointing, we introduce a customized check-pointing mode, \textbf{MiniMem}. It recomputes the $\mathrm{SO}(2)$ layer in the backward pass of \textit{FlashSO(2)}, trading speed for further memory savings.

\section{Results}
\label{sec:results}

We evaluate \textit{FlashSO(2)} at two levels. First, in \textbf{Sections}~\ref{subsec:equivariance} and \ref{subsec:roofline}, we consider the $\mathrm{SO}(2)$ layer on its own, where we measure the equivariance error of each precision setting and compare the native and fused implementations on a roofline model. In \textbf{Section}~\ref{subsec:evaluation}, we then integrate our \textit{FlashSO(2)} implementation into representative SO(2)-layer-based MLH (MALOQ \citep{maloq}) and MLIP (UMA \citep{uma}) models. Despite their shared $\mathrm{SO}(2)$ convolution architecture, these models differ in their number of layers, $l_{max}$, and $m_{max}$, as discussed in \textbf{Section}~\ref{subsec:mlh_and_mlip}. We report accuracy, training time, inference time, and peak memory. All experiments were run on a single NVIDIA GH200 superchip, except the training runs of \textbf{Fig.}~\ref{fig:maloq_results}(a).

\begin{figure}[t]
  \centering
\includegraphics[width=\linewidth]{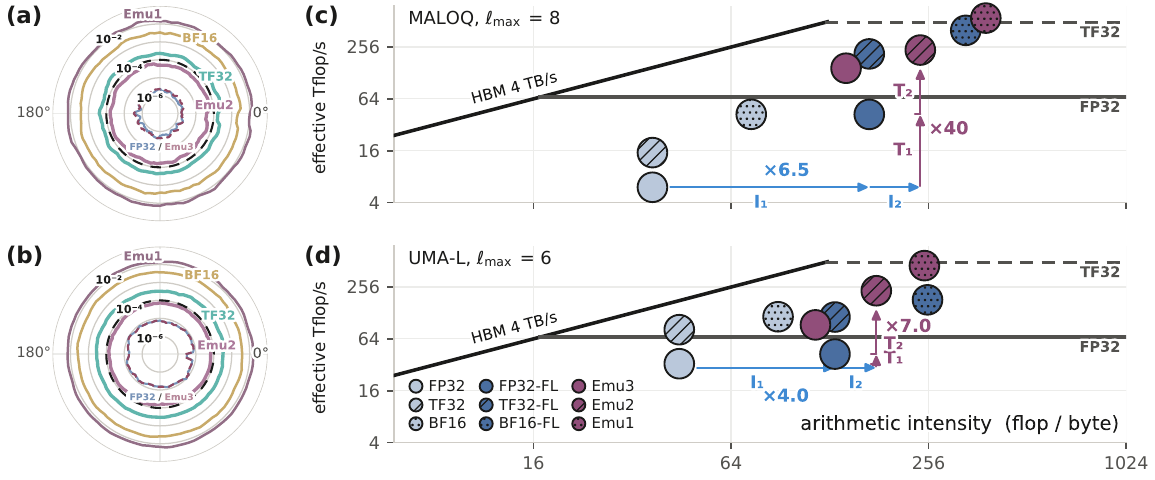}
  \caption{\textbf{Equivariance and roofline analysis of the $\mathrm{SO}(2)$ convolutions under low precision.}
    Rotation-equivariance residual of the (a) MALOQ Hamiltonian output head and (b) UMA forces against the rotation angle. The dashed ring marks the $10^{-4}$ tolerance.
    Roofline of the forward pass of (c) MALOQ and (d) UMA-Large (UMA-L) models. 
    Light blue symbols: native implementation; dark blue: addition of fused layer(-FL); purple: fused layer with INT8 emulation (Emu$1$--Emu$3$).
    The arrows split the gain over the native layer into kernel fusion ($I_1$ in intensity, $T_1$ in throughput) and INT8 emulation ($I_2$, $T_2$). $I_1$ = 4.6, $T_1$ = 7.0, $I_2$ = 1.4 and $T_2$ = 5.6 in (c), and 2.9, 1.3, 1.4 and 5.5 in (d).}
\label{fig:roofline}
\end{figure}

\subsection{Equivariance analysis}
\label{subsec:equivariance}

Equivariant models enforce rotational symmetries through exact mathematical constants (e.g., Clebsch-Gordan coefficients), thus any rounding of the intermediate activations appears as a rotation-dependent error in the output. We quantify this error in \textbf{Fig.}~\ref{fig:roofline}(a-b) through polar plots, where the magnitude of each point represents the deviation between the model performing a rotation and a mathematically exact rotation of the target, measured as the largest component-wise residual of the equivariance identity. \textbf{Figure}~\ref{fig:roofline}(a) is concerned with MALOQ (rotating a Hamiltonian target), \textbf{Fig.}~\ref{fig:roofline}(b) with UMA (rotating a force target). In both applications, the $\mathrm{SO}(2)$ layer is run at varying precisions, and an equivariance tolerance criterion of $10^{-4}$ (dashed ring) is adopted, one order of magnitude below the $10^{-3}$ value used by e3nn \citep{e3nn}. 

Among the native precision formats, only FP32 satisfies this criterion, explaining why symmetry-constrained models typically execute $\mathrm{SO}(2)$ layers on FP32 vector units. 
The native low-precision formats fail the equivariance test for both MALOQ and UMA: TF32's results lie just outside the threshold ring, whereas BF16's ones are two orders of magnitude further out, as our Emu$1$ approach. 
Emu$2$ instead stays inside the ring on both target models, and Emu$3$ is indistinguishable from FP32. Overall, Emu$2$ is the most computationally efficient setting on MMUs, while simultaneously fulfilling the tolerance criterion.

\subsection{Roofline models of the SO(2) layer}
\label{subsec:roofline}

We show roofline models of the native and \textit{FlashSO(2)} implementation of the $\mathrm{SO}(2)$ layer at varying precisions, considering the embedding layout used in either MALOQ (\textbf{Fig.~\ref{fig:roofline}}(c)) or UMA (\textbf{Fig.~\ref{fig:roofline}}(d)). In both cases, we select the computationally `heaviest' model settings. This corresponds to $l_{max}$ up to 8, $m_{max}=l_{max}$ in MALOQ, and UMA's `Large' (UMA-L) configuration where $l_{max}=6$, $m_{max}=2$. The arithmetic intensity quantifies the number of floating point operations per memory access. In both axes of \textbf{Fig.~\ref{fig:roofline}}(c-d), the number of floating point operations is counted according to FP32 arithmetic. An emulated run is therefore credited with the FP32 operations it replaces rather than the INT8 operations it executes. The arrows split the gain from the FP32 baseline to the Emu$2$ implementation into two steps, (1) kernel fusion ($I_1$, $T_1$) and (2) INT8 emulation ($I_2$, $T_2$).

\paragraph{Kernel Fusion} Starting from the reference implementation, which executes the pipeline of \textbf{Section}~\ref{subsec:pipeline} as a sequence of separate PyTorch kernels, we first isolate the performance improvements provided by kernel fusion (FL) alone, with no change in precision. 

In both \textbf{Fig.~\ref{fig:roofline}}(c) and \textbf{Fig.~\ref{fig:roofline}}(d), the reference implementation is shown as the lightest blue circles. In MALOQ (UMA-L) it reaches only 6 (33) effective Tflop/s with 37 (45) flop/byte arithmetic intensity, 9\% (49\%) of the FP32 vector peak and 4\% (18\%) of what the memory bandwidth allows at this intensity. The $\mathrm{SO}(2)$ layer is thus neither compute nor memory bound. 
The kernel fusion strategy from \textbf{Section}~\ref{subsec:flashso2} pushes the intermediate activations between the matrix products to stay in registers and shared memory. As a consequence, the arithmetic intensity rises by $I_1 = 4.6\times$ (MALOQ) and $2.9\times$ (UMA-L), the throughput by $T_1 = 7.0\times$ and $1.3\times$, respectively. Our fused FP32 layer reaches an overall 63\% of the FP32 vector peak performance in both models. At this point, the FP32 vector units are the new limit, and further gains must come from the MMUs.

\paragraph{INT8 Emulation} Starting from the fused $\mathrm{SO}(2)$ layer, we now push the matrix products onto the INT8 MMUs. We use the precision emulation (Emu$1$--Emu$3$) detailed in \textbf{Section}~\ref{subsec:emulation}.
Emulation also stores the activations as INT8 slices, with two bytes per element instead of four.

Doing this further increases the arithmetic intensity by $I_2$ = $1.4\times$, and throughput by $T_2$ = $5.6\times$ for MALOQ, while $I_2$ = $1.4\times$ and $T_2$ = $5.5\times$ for UMA-L, as compared to the case with fused kernels. In total Emu$2$ delivers $I_1\times I_2$ = $6.5\times$ the intensity and $T_1\times T_2$ = $40\times$ the throughput of the native $\mathrm{SO}(2)$ layer in MALOQ, $4.0\times$ and $7\times$ in UMA-L, respectively.
Hence, the Emu$2$ layer sustains 236 effective Tflop/s in MALOQ and 232 in UMA-L, $3.5\times$ the peak performance of the FP32 unit (67 Tflop/s on GH200 GPUs).  
Note that Emu$3$ achieves 145 Tflop/s in MALOQ, $2.2\times$ the FP32 peak, while reproducing FP32 down to its rounding floor.
For comparison, in \textbf{Fig.~\ref{fig:roofline}} we also show the performance of the fused $\mathrm{SO}(2)$ layer in native low-precision formats (TF32, BF16).
\textbf{Together, Emu$2$ delivers higher throughput and lower memory traffic than TF32 while staying within the equivariance tolerance, which no native low-precision format does.}

\subsection{Accelerating ML materials models}
\label{subsec:evaluation}

\begin{figure}[!t]
  \centering
  \includegraphics[width=\linewidth]{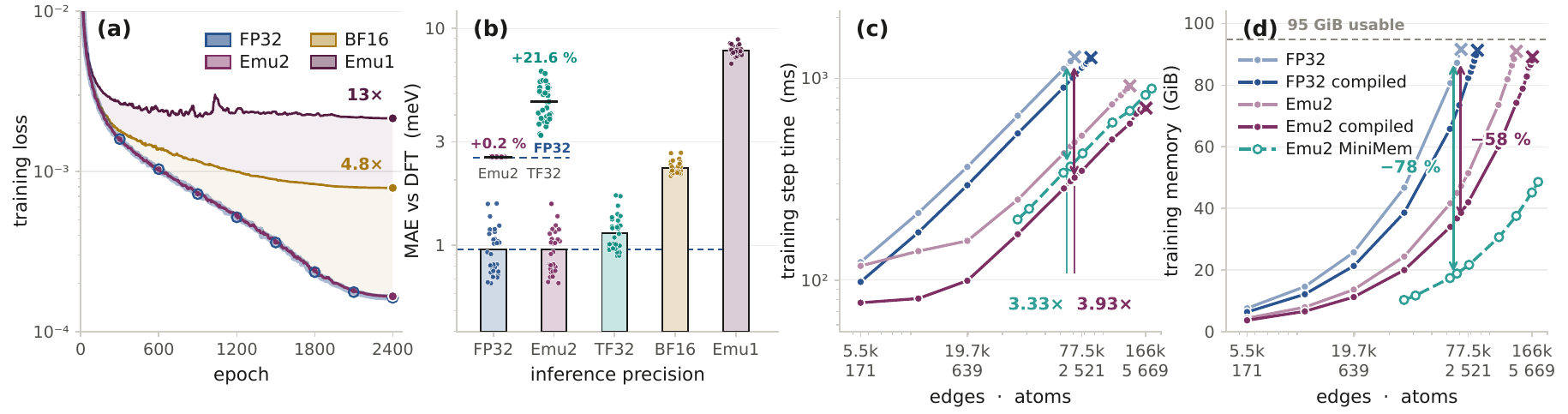}
  \caption{\textbf{Accelerating MALOQ's training.}
    (a)~Training loss curve with different precisions on the nablaDFT-2K dataset.
    (b)~Inference error against the 32 held-out conformers, weights are trained with Emu$2$ in (a). Inset: MAE of the Hamiltonian elements of each molecule relative to FP32.
    (c)~Training step time and (d)~peak memory as a function of the input size (number of atoms/edges).}
  \label{fig:maloq_results}
  \vspace{0.5\floatsep}
  \includegraphics[width=\linewidth]{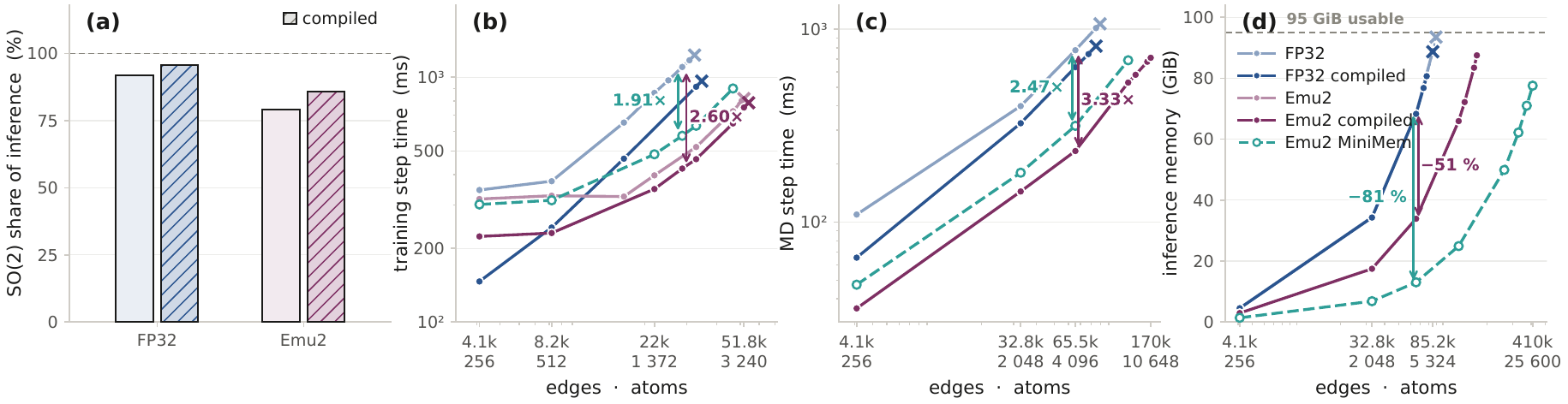}
  \caption{\textbf{Accelerating UMA-M's inference.}
    (a)~Fraction of the inference step spent in the $\mathrm{SO}(2)$ layer.
    (b)~Training step time, (c) MD step time, and (d) peak memory against the input size.}
  \label{fig:uma_results}
\end{figure}

Next, we analyze the performance of both MALOQ (\textbf{Fig.~\ref{fig:maloq_results}}) and the medium-model configuration of UMA, UMA-M (\textbf{Fig.~\ref{fig:uma_results}}) under real usage scenarios, highlighting the benefit of \textit{FlashSO(2)} over the baseline implementation of these models. For MALOQ, we focus on training, showing that \textit{FlashSO(2)} can efficiently tackle the densely connected graphs and high $l_{max}$ encountered by MLH models. We perform benchmarks with molecular subsets of $\nabla^2$DFT ($l_{max}=4$) \citep{nabla2dft}. For UMA-M, we focus on inference acceleration for latency-sensitive MD applications, on fcc-Cu supercells. In all benchmarks, the batch size is quantified by the number of edges, as this quantity determines the peak memory consumption. The number of atoms per batch depends on the structural distribution, and is provided in each case for reference. 

\paragraph{Accuracy evaluation}
First, we demonstrate that, in contrast to native low-precision formats, the accuracy of Emu$2$ is sufficient for both training and inference.
\textbf{Figure~\ref{fig:maloq_results}}(a) presents the training loss curve of MALOQ on the nablaDFT-2K subset of $\nabla^2$DFT, with the $\mathrm{SO}(2)$ layer kept at one precision for the whole run. We train all models to a fixed 2,400 epochs. At the end of training, the FP32 and Emu$2$ curves exactly coincide, while BF16 and Emu$1$ flatten out early and end at $4.8\times$ and $13\times$ the final loss of Emu$2$, respectively. 
We then take the weights trained in Emu$2$ and run inference at each precision on 32 held-out conformers (\textbf{Fig.~\ref{fig:maloq_results}}(b)). 
The mean absolute error (MAE) of the Hamiltonian elements against DFT is within 0.2\% of the FP32 value for Emu$2$, but 21.6\% higher with PyTorch's global TF32 mode enabled, and 2.5$\times$ (9.1$\times$) the FP32 value for BF16 (Emu$1$).
A model trained with Emu$2$ can thus be deployed in FP32 without any loss, whereas a TF32 fallback for inference will generally fail.

\paragraph{Training acceleration}
For MLH models, training is especially computationally intensive due to the highly connected graphs typically required to learn edge-level matrix targets \citep{maloq}. When treating large structures, large volumes of embeddings are created during MP iterations, leading to high memory consumption. \textit{FlashSO(2)} with Emu$2$ accelerates each of MALOQ's training steps by up to $3.93\times$ (\textbf{Fig.~\ref{fig:maloq_results}}(c)), with increasing gain as the input size (number of atoms/edges in the input molecular batch) grows. This speedup factor is obtained for the largest input that can be fit in memory in FP32 precision, under torch.compile. Since \textit{FlashSO(2)} keeps only INT8 slices between its kernels for the backward pass, Emu$2$ halves the peak memory. At the input size where FP32 runs out of memory, Emu$2$ lowers the peak training memory of MALOQ by 58\% under torch.compile. Switching to the `MiniMem' mode, \textit{FlashSO(2)}'s customized activation check-pointing implementation, trades part of the overall speedup for an even larger reduction in memory consumption of 78\%, while remaining $3.33\times$ faster than the FP32 reference. 

In UMA-M, where the $\mathrm{SO}(2)$ layer takes 79\% to 91\% of each training step (\textbf{Fig.~\ref{fig:uma_results}}(a)), adopting Emu$2$ accelerates each step by $2.6\times$, or $1.91\times$ with MiniMem (\textbf{Fig.~\ref{fig:uma_results}}(b)).

\paragraph{Inference acceleration}
Molecular dynamics (MD) simulations with MLIPs are latency-sensitive applications, where fast inference is crucial. Inference for an MD step involves a forward pass followed by force evaluations during the backward pass. Emu$2$ runs the UMA-M inference step $3.33\times$ faster than FP32 (\textbf{Fig.~\ref{fig:uma_results}}(c)). This factor was measured at 65.5k edges (4.1k atoms), the largest input that the FP32 run can fit into the memory of a single GPU, under torch.compile; MiniMem is $2.47\times$ faster than FP32 at the same structure size, while providing 81\% reduction in memory footprint (\textbf{Fig.}~\ref{fig:uma_results}(d)). Pure FP32 implementations exhaust the 95~GiB of the GH200 at 5.3k atoms in UMA-M inference. In comparison, MiniMem fits an MD step of 25.6k atoms (410k edges) in 78~GiB, vastly increasing the scale of MD simulations on a single GPU.

\section{Conclusion \& Outlook}

We present \textit{FlashSO(2)}, a customized, emulation-driven implementation of the $\mathrm{SO}(2)$ layer powering many of the top MLIP and MLH models in ML-accelerated computational materials science. 
Across the different models we evaluated, Emu$2$ improves over TF32, the most accurate native format of the GPU's MMUs, on three axes at once: (i) accuracy (same as FP32), (ii) speed (1.1$\times$ faster than TF32), and (iii) memory consumption (half of TF32), thus pushing the speed-accuracy Pareto frontier beyond native reduced-precision formats of MMUs. 

Along with enabling the exploitation of scaling laws at reduced computational cost, our implementation offers an avenue around several practical limitations faced by materials models today: (1) The use of short cutoff radii ($r_{cut}$) in MLIPs leads to known flaws in the representation of long-range interactions. Increasing $r_{cut}$ is challenging as the edge connectivity grows with the cube of this parameter. By decreasing the effective memory footprint by up to 81\%, Emu$2$ enables a $1.7\times$ increase in cutoff for the same structure size. 
(2) Through cross domain pretraining, the vast quantities of available electronic structure data can be leveraged to improve the prediction of energies or forces \citep{helm}. 
As shown in \textbf{Fig.}~\ref{fig:dataset_bubble_plot_gb}, generating a roughly 50\,GB DFT dataset of energies and forces produces more than 100\,TB of electronic structure data as intermediate products. So far, pretraining on these high-rank targets has been computationally heavy. 
We offer an avenue to take advantage of them at $40\times$ the throughput of reference. Code will be released upon publication.

\subsection*{AI use statement}
The author used Claude (Anthropic) to improve the grammar and readability of drafts written by the author and to help with LaTeX formatting. All technical content, results, and scientific claims are the authors’ own, and all LLM-assisted text was reviewed and verified by the authors.

\subsection*{Acknowledgments}
This work was supported by the Swiss National Science Foundation (SNSF) under grant n\textdegree~209358 (QuaTrEx), and by the Platform for Advanced Scientific Computing (PASC) in Switzerland (BoostQT). We acknowledge the scientific support and HPC resources from CSCS under projects c33, lp16, and lp82.

\bibliography{ref}
\bibliographystyle{iclr2027_conference}

\clearpage

\end{document}

%% file: math_commands.tex
\usepackage{amsmath,amsfonts,bm}

\def\eqref#1{equation~\ref{#1}}

\def\1{\bm{1}}

\DeclareMathAlphabet{\mathsfit}{\encodingdefault}{\sfdefault}{m}{sl}
\SetMathAlphabet{\mathsfit}{bold}{\encodingdefault}{\sfdefault}{bx}{n}



%% file: ref.bib
@inproceedings{wanet,
title={Enhancing the Scalability and Applicability of {Kohn-Sham} {Hamiltonians} for Molecular Systems},
author={Yunyang Li and Zaishuo Xia and Lin Huang and Xinran Wei and Samuel Harshe and Han Yang and Erpai Luo and Zun Wang and Jia Zhang and Chang Liu and Bin Shao and Mark Gerstein},
booktitle={The Thirteenth International Conference on Learning Representations},
year={2025},
url={https://openreview.net/forum?id=twEvvkQqPS}
}

@article{nequip,
  title = {E(3)-equivariant graph neural networks for data-efficient and accurate interatomic potentials},
  volume = {13},
  ISSN = {2041-1723},
  url = {http://dx.doi.org/10.1038/s41467-022-29939-5},
  DOI = {10.1038/s41467-022-29939-5},
  number = {1},
  journal = {Nature Communications},
  publisher = {Springer Science and Business Media LLC},
  author = {Batzner,  Simon and Musaelian,  Albert and Sun,  Lixin and Geiger,  Mario and Mailoa,  Jonathan P. and Kornbluth,  Mordechai and Molinari,  Nicola and Smidt,  Tess E. and Kozinsky,  Boris},
  year = {2022},
  month = May 
}

@misc{e3nn,
  doi = {10.48550/ARXIV.2207.09453},
  url = {https://doi.org/10.48550/ARXIV.2207.09453},
  author = {Geiger,  Mario and Smidt,  Tess},
  title = {e3nn: Euclidean Neural Networks},
  publisher = {arXiv},
  year = {2022},
  copyright = {Creative Commons Attribution 4.0 International}
}

@article{pet,
  title = {{PET-MAD} as a lightweight universal interatomic potential for advanced materials modeling},
  volume = {16},
  ISSN = {2041-1723},
  url = {http://dx.doi.org/10.1038/s41467-025-65662-7},
  DOI = {10.1038/s41467-025-65662-7},
  number = {1},
  journal = {Nature Communications},
  publisher = {Springer Science and Business Media LLC},
  author = {Mazitov,  Arslan and Bigi,  Filippo and Kellner,  Matthias and Pegolo,  Paolo and Tisi,  Davide and Fraux,  Guillaume and Pozdnyakov,  Sergey and Loche,  Philip and Ceriotti,  Michele},
  year = {2025},
  month = nov 
}

@misc{mace,
  doi = {10.48550/ARXIV.2206.07697},
  url = {https://arxiv.org/abs/2206.07697},
  author = {Batatia,  Ilyes and Kovács,  Dávid Péter and Simm,  Gregor N. C. and Ortner,  Christoph and Csányi,  Gábor},
  title = {{MACE}: Higher Order Equivariant Message Passing Neural Networks for Fast and Accurate Force Fields},
  publisher = {arXiv},
  year = {2022},
  copyright = {arXiv.org perpetual,  non-exclusive license}
}

@InProceedings{esen,
  title = 	 {Learning Smooth and Expressive Interatomic Potentials for Physical Property Prediction},
  author =       {Fu, Xiang and Wood, Brandon M and Barroso-Luque, Luis and Levine, Daniel S. and Gao, Meng and Dzamba, Misko and Zitnick, C. Lawrence},
  booktitle = 	 {Proceedings of the 42nd International Conference on Machine Learning},
  pages = 	 {17875--17893},
  year = 	 {2025},
  editor = 	 {Singh, Aarti and Fazel, Maryam and Hsu, Daniel and Lacoste-Julien, Simon and Berkenkamp, Felix and Maharaj, Tegan and Wagstaff, Kiri and Zhu, Jerry},
  volume = 	 {267},
  series = 	 {Proceedings of Machine Learning Research},
  month = 	 {Jul},
  publisher =    {PMLR},
  url = 	 {https://proceedings.mlr.press/v267/fu25h.html}
}

@inproceedings{uma,
title={{UMA}: A Family of Universal Models for Atoms},
author={Brandon M Wood and Misko Dzamba and Xiang Fu and Meng Gao and Muhammed Shuaibi and Luis Barroso-Luque and Kareem Abdelmaqsoud and Vahe Gharakhanyan and John R. Kitchin and Daniel S. Levine and Kyle Michel and Anuroop Sriram and Taco Cohen and Abhishek Das and Sushree Jagriti Sahoo and Ammar Rizvi and Zachary Ward Ulissi and C. Lawrence Zitnick},
booktitle={The Thirty-ninth Annual Conference on Neural Information Processing Systems},
year={2025},
url={https://openreview.net/forum?id=SvopaNxYWt}
}

@misc{hip,
  doi = {10.48550/ARXIV.2509.21624},
  url = {https://doi.org/10.48550/ARXIV.2509.21624},
  author = {Burger,  Andreas and Thiede,  Luca and Rønne,  Nikolaj and Bernales,  Varinia and Vijaykumar,  Nandita and Vegge,  Tejs and Bhowmik,  Arghya and Aspuru-Guzik,  Alan},
  title = {Shoot from the HIP: Hessian Interatomic Potentials without derivatives},
  publisher = {arXiv},
  year = {2025},
  copyright = {Creative Commons Attribution Non Commercial Share Alike 4.0 International}
}

@misc{qhflow,
  doi = {10.48550/ARXIV.2505.18817},
  url = {https://arxiv.org/abs/2505.18817},
  author = {Kim,  Seongsu and Kim,  Nayoung and Kim,  Dongwoo and Ahn,  Sungsoo},
  title = {High-order Equivariant Flow Matching for Density Functional Theory {Hamiltonian} Prediction},
  publisher = {arXiv},
  year = {2025},
  copyright = {Creative Commons Attribution 4.0 International}
}

@misc{emugemm,
  doi = {10.48550/ARXIV.2606.25453},
  url = {https://arxiv.org/abs/2606.25453},
  author = {Lu,  Denghui and Maeder,  Alexander and Luisier,  Mathieu and Ziogas,  Alexandros Nikolaos},
  title = {{EmuGEMM}: Fused Tensor Core Kernels for Precision Emulation in Matrix Multiplication},
  publisher = {arXiv},
  year = {2026},
  copyright = {arXiv.org perpetual,  non-exclusive license}
}

@misc{deeph_pack,
  doi = {10.48550/ARXIV.2601.02938},
  url = {https://arxiv.org/abs/2601.02938},
  author = {Li,  Yang and Wang,  Yanzhen and Zhao,  Boheng and Gong,  Xiaoxun and Wang,  Yuxiang and Tang,  Zechen and Wang,  Zixu and Yuan,  Zilong and Li,  Jialin and Sun,  Minghui and Chen,  Zezhou and Tao,  Honggeng and Wu,  Baochun and Yu,  Yuhang and Li,  He and da Jornada,  Felipe H. and Duan,  Wenhui and Xu,  Yong},
  title = {{DeepH-pack}: A general-purpose neural network package for deep-learning electronic structure calculations},
  publisher = {arXiv},
  year = {2026},
  copyright = {arXiv.org perpetual,  non-exclusive license}
}

@misc{helm,
  doi = {10.48550/ARXIV.2510.00224},
  url = {https://arxiv.org/abs/2510.00224},
  author = {Kaniselvan,  Manasa and Miller,  Benjamin Kurt and Gao,  Meng and Nam,  Juno and Levine,  Daniel S.},
  title = {Learning from the electronic structure of molecules across the periodic table},
  publisher = {arXiv},
  year = {2025},
  copyright = {arXiv.org perpetual,  non-exclusive license}
}

@misc{esen_big,
  doi = {10.48550/ARXIV.2609.09293},
  url = {https://arxiv.org/abs/2609.09293},
  author = {Gao,  Meng and Zadeh,  Armin Shayesteh and Seal,  Aniruddha and Dasetty,  Siva and Achar,  Siddarth K. and Dzamba,  Misko and Miller,  Benjamin K. and Jacobson,  Leif D. and Zitnick,  C. Lawrence and Wood,  Brandon M. and Ulissi,  Zachary W. and Levine,  Daniel S. and Ferguson,  Andrew L.},
  title = {Quantum-accurate atomistic modeling of enzyme catalysis using a machine learned potential},
  publisher = {arXiv},
  year = {2026},
  copyright = {Creative Commons Attribution 4.0 International}
}

@misc{equiformerv2,
  doi = {10.48550/ARXIV.2306.12059},
  url = {https://arxiv.org/abs/2306.12059},
  author = {Liao,  Yi-Lun and Wood,  Brandon and Das,  Abhishek and Smidt,  Tess},
  title = {{EquiformerV2}: Improved Equivariant Transformer for Scaling to Higher-Degree Representations},
  publisher = {arXiv},
  year = {2023},
  copyright = {Creative Commons Attribution 4.0 International}
}

@article{hamgnn,
  title = {Transferable equivariant graph neural networks for the {Hamiltonians} of molecules and solids},
  volume = {9},
  ISSN = {2057-3960},
  url = {http://dx.doi.org/10.1038/s41524-023-01130-4},
  DOI = {10.1038/s41524-023-01130-4},
  number = {1},
  journal = {npj Computational Materials},
  publisher = {Springer Science and Business Media LLC},
  author = {Zhong,  Yang and Yu,  Hongyu and Su,  Mao and Gong,  Xingao and Xiang,  Hongjun},
  year = {2023},
  month = Oct 
}

@misc{dpa4,
  doi = {10.48550/ARXIV.2606.02419},
  url = {https://arxiv.org/abs/2606.02419},
  author = {Li,  Tiancheng and Li,  Wentao and Peng,  Anyang and Xue,  Jianming and Zhang,  Linfeng and Zhang,  Duo and Wang,  Han},
  title = {{DPA4}: Pushing the Accuracy-Cost Frontier of Interatomic Potentials with {EMFA} {SO(2)} Convolution},
  publisher = {arXiv},
  year = {2026},
  copyright = {Creative Commons Attribution 4.0 International}
}

@misc{mixed_precision,
  doi = {10.48550/ARXIV.1710.03740},
  url = {https://arxiv.org/abs/1710.03740},
  author = {Micikevicius,  Paulius and Narang,  Sharan and Alben,  Jonah and Diamos,  Gregory and Elsen,  Erich and Garcia,  David and Ginsburg,  Boris and Houston,  Michael and Kuchaiev,  Oleksii and Venkatesh,  Ganesh and Wu,  Hao},
  title = {Mixed Precision Training},
  publisher = {arXiv},
  year = {2017},
  copyright = {arXiv.org perpetual,  non-exclusive license}
}

@misc{the_quantization_paper,
  doi = {10.48550/ARXIV.1712.05877},
  url = {https://arxiv.org/abs/1712.05877},
  author = {Jacob,  Benoit and Kligys,  Skirmantas and Chen,  Bo and Zhu,  Menglong and Tang,  Matthew and Howard,  Andrew and Adam,  Hartwig and Kalenichenko,  Dmitry},
  title = {Quantization and Training of Neural Networks for Efficient Integer-Arithmetic-Only Inference},
  publisher = {arXiv},
  year = {2017},
  copyright = {arXiv.org perpetual,  non-exclusive license}
}

@InProceedings{so2,
  title = 	 {Reducing {SO}(3) Convolutions to {SO}(2) for Efficient Equivariant {GNN}s},
  author =       {Passaro, Saro and Zitnick, C. Lawrence},
  booktitle = 	 {Proceedings of the 40th International Conference on Machine Learning},
  pages = 	 {27420--27438},
  year = 	 {2023},
  editor = 	 {Krause, Andreas and Brunskill, Emma and Cho, Kyunghyun and Engelhardt, Barbara and Sabato, Sivan and Scarlett, Jonathan},
  volume = 	 {202},
  series = 	 {Proceedings of Machine Learning Research},
  month = 	 {Jul},
  publisher =    {PMLR},
  url = 	 {https://proceedings.mlr.press/v202/passaro23a.html}
}

@article{ozaki2012,
  title = {Error-free transformations of matrix multiplication by using fast routines of matrix multiplication and its applications},
  volume = {59},
  ISSN = {1572-9265},
  url = {http://dx.doi.org/10.1007/s11075-011-9478-1},
  DOI = {10.1007/s11075-011-9478-1},
  number = {1},
  journal = {Numerical Algorithms},
  publisher = {Springer Science and Business Media LLC},
  author = {Ozaki, Katsuhisa and Ogita, Takeshi and Oishi, Shin'ichi and Rump, Siegfried M.},
  year = {2012},
  month = jan,
  pages = {95--118}
}

@article{ootomo2024,
  title = {{DGEMM} on integer matrix multiplication unit},
  volume = {38},
  ISSN = {1741-2846},
  url = {http://dx.doi.org/10.1177/10943420241239588},
  DOI = {10.1177/10943420241239588},
  number = {4},
  journal = {The International Journal of High Performance Computing Applications},
  publisher = {SAGE Publications},
  author = {Ootomo, Hiroyuki and Ozaki, Katsuhisa and Yokota, Rio},
  year = {2024},
  month = mar,
  pages = {297--313}
}

@misc{maloq,
  doi = {10.48550/ARXIV.2606.28911},
  url = {https://arxiv.org/abs/2606.28911},
  author = {Kaniselvan,  Manasa and Maeder,  Alexander and Lu,  Denghui and Ziogas,  Alexandros Nikolaos and Luisier,  Mathieu},
  title = {{MALOQ}: Massively Accelerated Learning of Operators for Quantum Transport},
  publisher = {arXiv},
  year = {2026},
  copyright = {arXiv.org perpetual,  non-exclusive license}
}

@misc{fp8,
  doi = {10.48550/ARXIV.2209.05433},
  url = {https://arxiv.org/abs/2209.05433},
  author = {Micikevicius,  Paulius and Stosic,  Dusan and Burgess,  Neil and Cornea,  Marius and Dubey,  Pradeep and Grisenthwaite,  Richard and Ha,  Sangwon and Heinecke,  Alexander and Judd,  Patrick and Kamalu,  John and Mellempudi,  Naveen and Oberman,  Stuart and Shoeybi,  Mohammad and Siu,  Michael and Wu,  Hao},
  title = {{FP8} Formats for Deep Learning},
  publisher = {arXiv},
  year = {2022},
  copyright = {arXiv.org perpetual,  non-exclusive license}
}

@misc{deepseekv3,
  doi = {10.48550/ARXIV.2412.19437},
  url = {https://arxiv.org/abs/2412.19437},
  author = {{DeepSeek-AI}},
  title = {{DeepSeek-V3} Technical Report},
  publisher = {arXiv},
  year = {2024},
  copyright = {Creative Commons Attribution 4.0 International}
}

@inproceedings{fa3,
title={{FlashAttention-3}: Fast and Accurate Attention with Asynchrony and Low-precision},
author={Jay Shah and Ganesh Bikshandi and Ying Zhang and Vijay Thakkar and Pradeep Ramani and Tri Dao},
booktitle={The Thirty-eighth Annual Conference on Neural Information Processing Systems},
year={2024},
url={https://proceedings.neurips.cc/paper_files/paper/2024/hash/7ede97c3e082c6df10a8d6103a2eebd2-Abstract-Conference.html}
}

@misc{omol25,
  doi = {10.48550/ARXIV.2505.08762},
  url = {https://arxiv.org/abs/2505.08762},
  author = {Levine,  Daniel S. and Shuaibi,  Muhammed and Spotte-Smith,  Evan Walter Clark and Taylor,  Michael G. and Hasyim,  Muhammad R. and Michel,  Kyle and Batatia,  Ilyes and Csányi,  Gábor and Dzamba,  Misko and Eastman,  Peter and Frey,  Nathan C. and Fu,  Xiang and Gharakhanyan,  Vahe and Krishnapriyan,  Aditi S. and Rackers,  Joshua A. and Raja,  Sanjeev and Rizvi,  Ammar and Rosen,  Andrew S. and Ulissi,  Zachary and Vargas,  Santiago and Zitnick,  C. Lawrence and Blau,  Samuel M. and Wood,  Brandon M.},
  title = {The Open Molecules 2025 ({OMol25}) Dataset,  Evaluations,  and Models},
  publisher = {arXiv},
  year = {2025},
  copyright = {arXiv.org perpetual,  non-exclusive license}
}

@article{schnorb,
  title = {Unifying machine learning and quantum chemistry with a deep neural network for molecular wavefunctions},
  volume = {10},
  ISSN = {2041-1723},
  url = {http://dx.doi.org/10.1038/s41467-019-12875-2},
  DOI = {10.1038/s41467-019-12875-2},
  number = {1},
  journal = {Nature Communications},
  publisher = {Springer Science and Business Media LLC},
  author = {Schütt,  K. T. and Gastegger,  M. and Tkatchenko,  A. and Müller,  K.-R. and Maurer,  R. J.},
  year = {2019},
  month = nov
}

@inproceedings{qh9,
title={{QH9}: A Quantum {Hamiltonian} Prediction Benchmark for {QM9} Molecules},
author={Haiyang Yu and Meng Liu and Youzhi Luo and Alex Strasser and Xiaofeng Qian and Xiaoning Qian and Shuiwang Ji},
booktitle={Thirty-seventh Conference on Neural Information Processing Systems Datasets and Benchmarks Track},
year={2023},
url={https://arxiv.org/abs/2306.09549}
}

@inproceedings{nabla2dft,
title={{$\nabla^2$DFT}: A Universal Quantum Chemistry Dataset of Drug-Like Molecules and a Benchmark for Neural Network Potentials},
author={Kuzma Khrabrov and Anton Ber and Artem Tsypin and Konstantin Ushenin and Egor Rumiantsev and Alexander Telepov and Dmitry Protasov and Ilya Shenbin and Anton Alekseev and Mikhail Shirokikh and Sergey Nikolenko and Elena Tutubalina and Artur Kadurin},
booktitle={The Thirty-eighth Conference on Neural Information Processing Systems Datasets and Benchmarks Track},
year={2024},
url={https://arxiv.org/abs/2406.14347}
}

@misc{equiformerv3,
  doi = {10.48550/ARXIV.2604.09130},
  url = {https://arxiv.org/abs/2604.09130},
  author = {Liao,  Yi-Lun and Hoffman,  Alexander J. and Shen,  Sabrina C. and Duval,  Alexandre and Norwood,  Sam Walton and Smidt,  Tess},
  title = {{EquiformerV3}: Scaling Efficient,  Expressive,  and General {SE(3)}-Equivariant Graph Attention Transformers},
  publisher = {arXiv},
  year = {2026},
  copyright = {Creative Commons Attribution 4.0 International}
}

@inproceedings{cublas_emulation,
  title = {Guaranteed {DGEMM} Accuracy While Using Reduced Precision Tensor Cores Through Extensions of the {Ozaki} Scheme},
  author = {Schwarz, Angelika and Anders, Anton and Brower, Cole and Bayraktar, Harun and Gunnels, John and Clark, Kate and Xu, RuQing G. and Rodriguez, Samuel and Cayrols, Sebastien and Tabaszewski, Pawe{\l} and Podlozhnyuk, Victor},
  booktitle = {Proceedings of the Supercomputing Asia and International Conference on High Performance Computing in Asia Pacific Region (SCA/HPCAsia 2026)},
  pages = {91--101},
  publisher = {ACM},
  year = {2026},
  DOI = {10.1145/3773656.3773670}
}

@InProceedings{flashtp,
  title = 	 {{F}lash{TP}: Fused, Sparsity-Aware Tensor Product for Machine Learning Interatomic Potentials},
  author =       {Lee, Seung Yul and Kim, Hojoon and Park, Yutack and Jeong, Dawoon and Han, Seungwu and Park, Yeonhong and Lee, Jae W.},
  booktitle = 	 {Proceedings of the 42nd International Conference on Machine Learning},
  pages = 	 {33143--33156},
  year = 	 {2025},
  editor = 	 {Singh, Aarti and Fazel, Maryam and Hsu, Daniel and Lacoste-Julien, Simon and Berkenkamp, Felix and Maharaj, Tegan and Wagstaff, Kiri and Zhu, Jerry},
  volume = 	 {267},
  series = 	 {Proceedings of Machine Learning Research},
  month = 	 {Jul},
  publisher =    {PMLR},
  url = 	 {https://proceedings.mlr.press/v267/lee25l.html}
}

@inproceedings{openequivariance,
  title = {An Efficient Sparse Kernel Generator for {O(3)}-Equivariant Deep Networks},
  author = {Bharadwaj, Vivek and Glover, Austin and Bulu{\c{c}}, Ayd{\i}n and Demmel, James},
  booktitle = {2025 Proceedings of the Conference on Applied and Computational Discrete Algorithms (ACDA)},
  pages = {32--46},
  publisher = {Society for Industrial and Applied Mathematics},
  year = {2025},
  DOI = {10.1137/1.9781611978759.3}
}

@misc{cuequivariance,
  title = {{cuEquivariance}},
  author = {{NVIDIA}},
  url = {https://github.com/NVIDIA/cuEquivariance},
  year = {2024}
}

@inproceedings{ansel2024pytorch2,
  title={{PyTorch 2}: Faster Machine Learning Through Dynamic {Python} Bytecode Transformation and Graph Compilation},
  author={Ansel, Jason and Yang, Edward and He, Horace and Gimelshein, Natalia and Jain, Animesh and Voznesensky, Michael and Bao, Bin and Bell, Peter and Berard, David and Burovski, Evgeni and others},
  booktitle={Proceedings of the 29th ACM International Conference on Architectural Support for Programming Languages and Operating Systems, Volume 2 (ASPLOS '24)},
  pages={929--947},
  year={2024},
  doi={10.1145/3620665.3640366}
}
